\documentclass[letterpaper, 10 pt, conference]{ieeeconf}  %

\IEEEoverridecommandlockouts                              %

\usepackage{cite}
\usepackage{amsmath,amssymb,amsfonts}
\usepackage{algorithmic}
\usepackage{graphicx}
\usepackage{textcomp}
\usepackage{xcolor}
\usepackage{url}
\usepackage{gensymb}
\usepackage{siunitx}
\usepackage{censor}
\usepackage{lipsum}
\usepackage{balance}
\usepackage[ruled,vlined]{algorithm2e}

\graphicspath{{./images}}

\title{\LARGE \bf
Acoustic Ellipses: Bio-Inspired Omnidirectional Echolocation in Cooperative Multi-Agent Systems using Frequency Sweeps*
}

\author{{Petras Swissler}$^{1}$ and {Lindsay Burke}$^{2}$ and {Julia Hyland Bruno}$^{3}$%
\thanks{*{This work was supported by the New Jersey Institute of Technology}}%
\thanks{$^{1}$Petras Swissler is with Faculty of Mechanical and Industrial Engineering, New Jersey Institute of Technology, Newark, NJ 07102, USA
        {\tt\small petras.swissler@njit.edu}}%
\thanks{$^{2}$Lindsay Burke is with the Department of Robotics,
        University of Michigan, Ann Arbor, MI 48109, USA
        {\tt\small lrb@njit.edu}}%
\thanks{$^{2}$Julia Hyland Bruno is with the Department of Biological Sciences,
        New Jersey Institute of Technology, Newark, NJ 07102, USA
        {\tt\small julia.hylandbruno@njit.edu}}%
}

\usepackage{tikz}

\newcommand\copyrighttext{%
  \footnotesize © 2026 IEEE. Personal use of this material is permitted. Permission from IEEE must be obtained for all other uses, in any current or future media, including reprinting/republishing this material for advertising or promotional purposes, creating new collective works, for resale or redistribution to servers or lists, or reuse of any copyrighted component of this work in other works.}

\newcommand\copyrightnotice{%
  \begin{tikzpicture}[remember picture,overlay]
    \node[anchor=south,yshift=10pt] at (current page.south) {\fbox{\parbox{\dimexpr\textwidth-\fboxsep-\fboxrule\relax}{\copyrighttext}}};
  \end{tikzpicture}%
}

\begin{document}

\maketitle

\copyrightnotice

\thispagestyle{empty}
\pagestyle{empty}

\begin{abstract}
Inspired by the flight and song of birds, we propose an approach that enables cooperating agents to effectively identify obstacles in their environment by having a stationary source agent emit a frequency-modulated chirp while one or more listener agents observe the direct and reflected signals while in motion. We present a novel mapping approach that exploits the ``frequency gap'' between direct and reflected chirp signals to define candidate reflection ellipses, which are then fed into a 2D accumulation filter to identify locations with the highest density of potential reflections.

We demonstrate this work first with simulation results derived from an efficient, bespoke audio simulator, examining the effect of obstacle count, sampling rate, and path curvature on the ability to accurately identify environmental obstacles for a one-listener scenario. We then examine different cooperative motion strategies for two-listener configurations. Finally, we validate the real-world viability of this approach through field experiments in an outdoor park setting to demonstrate the ability to identify frequency gaps using off-the-shelf hardware. Our results provide a foundation for a low-cost approach to environmental mapping in swarm robotic systems.
\end{abstract}

\section{Introduction}

\begin{figure}
    \centering
    \includegraphics[width=0.9\columnwidth]{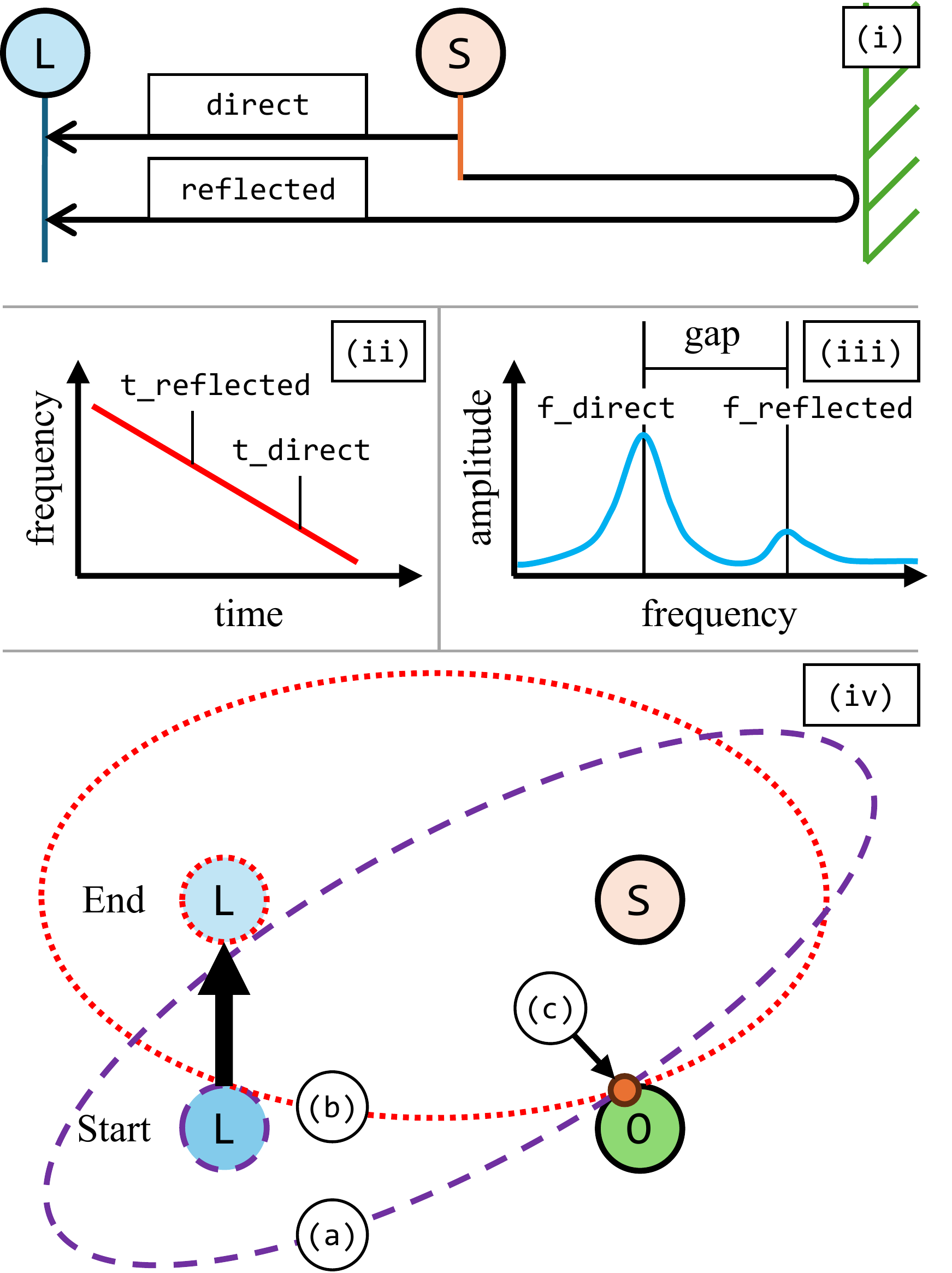}
    \caption{(i) Sound from the source (S) can travel to the listener (L) either via direct path or via a reflected path. (ii) If the sound is a frequency sweep, this will result in different frequencies being heard. (iii) The gap between frequencies is proportional to the difference in travel distance. (iv) Differences in sound travel distances can be represented as ellipses of possible reflection points (a and b). By sampling at multiple locations (such as when in motion) makes the ellipses tend to intersect around reflection points on obstacles (c).}
    \label{fig:concept}
\end{figure}

Biosonar systems such as bat and dolphin echolocation demonstrate that the reverberations of acoustic signals provide rich information for localizing objects and navigating in space \cite{moss2023adaptive, griffin1944echolocation}. Broadband signals in particular play an important role in biological range finding. Echolocating bats, for example, rely on the returning echoes of frequency-modulated (FM) calls for accurate target localization and classification \cite{schnitzler2001echolocation}. Biosonar is typically high-amplitude, beam-like, and ultrasonic. However, frequency sweeps are a common feature of animal vocalizations, from the howls of wolves \cite{hennelly2017howl}, to the ``whines" of frogs \cite{ryan1988constraints}, to the songs of whales \cite{clark2022baleen,mercado2018sonar} and, perhaps most notably, birds \cite{stowell2014large,goller2021different}. Why are chirps and other FM signals so pervasive? While bioacoustic research often focuses on decoding the possible information transferred between senders and receivers (e.g., about sex, status, behavioral state, imminent threats, etc.) \cite{kershenbaum2016disentangling}, it is possible that vocalizing and eavesdropping animals use sound to perceive their environments in dynamic ways that have been underappreciated. A few studies have shown that chirps contain built-in cues for ranging \cite{morton1996songbirds, mercado2007acoustic, mercado2017chickadee}. However, the extent to which birds and other animals use such cues to coordinate their behaviors is not well understood.

Robotics uses many sensor types for localization and mapping in uncertain environments, such as LiDAR and ultrasonic sensors. However, LiDAR is often expensive \cite{tabir_lidar}, especially in the context of large-scale robot swarms where this cost scales with the number of agents. Ultrasonic sensing is limited to short-range applications due to the rapid attenuation of high-frequency sound waves and has a minimum detection range due to continued vibrations in the transducer after a pulse is sent out \cite{mi13040520}. This presents the need for a low-cost range finding method, that is viable beyond short-range distance detection, for applications to real-time environment mapping in large-scale swarm systems. Additionally, using these sensors in a swarm is difficult due to the probability of cross-talk. We propose that off-the-shelf speakers and microphones can fill this role in cooperative robotic systems.

 In this paper, we consider the possibility that frequency-modulated signals could be used to enable cooperative environmental sensing of two robots that know each other's position (e.g, via GPS and via systems such as the Cricket indoor localization system \cite{priyantha2005cricket}), perhaps using ranging mechanisms similar to those used by biological systems. However, it is important to note that the way in which animals perceive sound differs greatly from artificial systems such as robots. With few exceptions, robots use typical microphones, which consist of a single thin film whose vibration induces small voltages in a coil \cite{eargle2012microphone}. These voltages are sampled at a very high frequency (e.g., \qty{44.1}{kHz} in the case of CD-quality music recordings \cite{pras2010sampling}) with precise timings. Interpretation of this signal then relies on algorithms such as the Fast Fourier Transform (FFT) to extract the frequencies present in the signal. In contrast, animal ears detect sound via the mechanosensory stereocilia of hair cells within the inner ear (in vertebrates) that are each maximally stimulated by different frequencies and thus transmit frequency-specific electrical signals to the brain \cite{altoe2017model}. In other words, much like the techniques used for ultrasonic sensing, robots rely on timing-based perception, whereas biological sensing is frequency-based. This difference drives the core question of this paper: What would a frequency-oriented approach to range sensing look like?%

The primary contribution of this paper is a novel approach to environmental mapping that uses auditory frequency sweeps to identify objects in the environment. We demonstrate the use of this algorithm in a 2D simulation and examine how different environmental characteristics impact its performance. We also demonstrate the use of this approach in several real-world experiments. Simulation code used towards this publication is available under a CC BY-NC-SA 4.0 license at \cite{github_anon}. 

Unlike traditional click-based acoustic positioning methods that must strictly control emission timings to avoid pulse overlap, the continuous frequency-gap approach we present in this paper entirely decouples measurement timing from signal emission, meaning that multiple passive listener agents can simultaneously use a single continuous source chirp without  channel congestion or multi-robot cross-talk.

\section{Principles and Methods}
\subsection{Conceptual Basis}

The use of auditory ``clicks'' for range finding is a common technique in robotics to identify environmental features \cite{moreno2002genetic}, with HC-SR04 modules being a common sensor used in hobby electronics \cite{d03n3rfr1tz3_hcsr04}. However, these click-based approaches are limited to single, instantaneous measurements, and there is a limit to %
how often clicks can be emitted since clicks are indistinguishable from each other.

Relative robot-to-robot (and device-to-device) localization and range finding has been demonstrated with simple sensors \cite{rothermich2004distributed, rubenstein2012kilobot, zhuang2016smartphone, ruiz2017comparing} or via networking and standard GPS, so mechanisms for robots to localize against peer robots is fairly well understood. What is less understood is how such multi-robot auditory range finding could be used for the purposes of collective mapping to replace more-expensive LiDAR and camera-based approaches.

We propose that environmental mapping can be achieved via audio signals by having one ``Listener'' robot move relative to another, stationary, ``Source'' robot while the Source emits a frequency sweep, also known as a ``chirp.'' Chirps are inherently continuous processes, and so the Listener will simultaneously hear sound that takes a direct path from the Source to the Listener as well as sound that reflects off of an obstacle, taking a longer path, as illustrated in Fig. \ref{fig:concept}.i. Though heard simultaneously, the sound will have originated at two separate times in the past, $t_{direct}$ and $t_{reflected}$, with $t_{reflected}$ occurring further in the past due to the time it takes for sound to travel (Fig. \ref{fig:concept}.ii). Assuming a monotonic frequency sweep (we use a linear sweep in this paper), this time difference corresponds to a difference in the frequencies that the Listener hears, and the gap between the direct-path frequency and reflected-path frequencies can be found (Fig. \ref{fig:concept}.iii). Note that this gap will be constant regardless of the exact time that the frequencies are analyzed, meaning that the exact times of the measurements do not matter: this frequency-gap formulation entirely decouples measurement timing from signal emission. In other words, the difference in heard frequencies corresponds to a calculable difference in the distances that the sounds have traveled, as we describe below.

Let $f_{start}$ be the start frequency of the chirp, $f_{end}$ be the end frequency of the chirp, $T$ be the duration of the chirp, $t_{start}$ be the time that the chirp begins, and $t$ be the current time. For the examples in this paper, the frequency output by the Source at any given time, $f_{output}$ is given as
\begin{equation}
    f_{output}(t) = 
f_{end} + (f_{start}-f_{end})\frac{t - t_{start}}{T} 
\end{equation}

for the domain $t_{start} \le t \le (t_{start} + T)$. Please note that this chirp formulation was chosen for simplicity of analysis, but, with modifications to the approach prented in this paper, could be any monotonic relationship between $t$ and $f_{output}$.

Using this known equation for a chirp, any observed $\Delta f$ can be translated to a time difference $\Delta t$ using

\begin{equation}
    \Delta t = \frac{\Delta f \cdot T}{f_{start}-f_{end}}.
\end{equation}

Using the speed of sound $a$, the difference in traveled distance, $\Delta d$ is calculated as 

\begin{equation}
    \Delta d = a \cdot \Delta t.
\end{equation}

We assume that we know the location of the Source and the Listener through some traditional global positioning system, and that the distance between them is calculated to be $D$. The position of the Listener and the position of the Source are the foci of an ellipse with a major axis of length $D+\Delta d$. 

To explain why this results in an ellipse, consider the diagram in Fig. \ref{fig:concept}.i. The direct distance between the Source and Listener is $D$. The only constraint on the reflected distance is that $d_{reflected} = D + \Delta d$, since the $\Delta t$ derived from the difference in observed frequencies only accounts for the difference in sound travel distance, and so the known distance $D$ must be added to obtain the total $d_{reflected}$. A useful property of ellipses is that that all points on an ellipse have a constant sum of distances to the two foci, and this sum of distances is equal to the length of the major axis. The diagram in Fig. \ref{fig:concept}.i shows the simplest case where the Listener, Source, and reflection are collinear. However, as illustrated in Fig. \ref{fig:concept}.iv, the point of reflection could be anywhere on an ellipse satisfying the equation

\begin{eqnarray}
    \nonumber \sqrt{(x-x_{Source})^2+(y-y_{Source})^2} +\\
    \sqrt{(x-x_{Listener})^2+(y-y_{Listener})^2} = D+ \Delta d.
\end{eqnarray}
\vspace{0.001mm}

This approach has parallels to Frequency Modulated Continuous Wave (FMCW) approaches to radar \cite{stove1992linear}. However, the proposed method is distinct from such approaches in several ways. Most FMCW systems are monostatic, whereas our approach has the Listener and Source as separate entities, where the Listener moves independently of the static Source. This difference enables us to consider the full range of possible reflection points (the candidate ellipses), rather than the 1D range measurements of FMCW systems. FMCW approaches rely on beat frequency analysis, rather than directly comparing frequencies, meaning that such systems must inherently operate using high frequency signals. Our addition of movement (sampling multiple Listener locations) to extract more information from the environment has parallels with virtual aperture approaches to sonar and radar \cite{marx2000introduction}, though most such approaches rely on timing-based perception rather than frequency-gap perception.

\subsection{2D Particle-Based Audio Simulator}

We created a bespoke 2D audio simulator  in Matlab. In this simulator, sound is modeled as discrete particles having a position, velocity, frequency, and loudness properties. These sound particles are emitted in all directions from a Source, with 3600 particles being emitted every time step for the simulations presented in this paper. Particles are evenly distributed in a circle with velocities pointing away from the center of the Source; a random angular offset is applied to help prevent dead zones far away from the Source. 

These particles are able to reflect off of obstacles in the environment. Although arbitrary polygonal obstacles are supported, obstacles for the tests presented here are modeled as perfect circles. In addition to their shape, obstacles have a roughness and absorption value. Roughness defines the amount of randomness applied to a sound particle bouncing off of it. A value of 10\% is used for simulations presented in this paper. Absorption defines the effect of reflections on the loudness of the sound particle. A 20\% absorption is used for results in this paper. These numbers are not based on a specific material, but are in line with common materials such as plywood and brick \cite{everest_pohlmann_mha_appendix, Zeng2006Scattering}

The particles are eventually detected by a Listener which senses all particles within some radius of its position. A larger radius introduces difficulties in determining exact distances but a smaller radius reduces the number of audio particles that it observes, which also reduces the accuracy of the system due to fewer sound particles being sensed at any given time. %
A spectrogram is created by assigning each sensed sound particle to a bin according to its frequency and summing the loudness of each particle within that bin. Peaks are then identified and gaps between peaks are calculated.

This approach to simulating sound is reasonably performant and made it possible to easily try different approaches to the localization problem. The primary shortcoming of using particles rather than modeling actual sound waves as in high-fidelity simulators such as \cite{scheibler2018pyroomacoustics, odeon19, cadnaa} is the absence of phenomena such frequency-dependent noise absorption  and sound diffusion through materials, both of which would not affect our approach. %

\subsection{Implementation of Algorithm in the Simulator}

In the code, ellipses are created by first generating points on a horizontal ellipse, rotating it by the appropriate amount, then shifting those points to the appropriate location. These points are placed into the bins of a discretized 2D accumulation filter, which are sized to represent squares with \qty{0.1}{m} sides. As the Listener moves, the candidate ellipse will change, however it will tend to have a greater number of intersections on the surface of the obstacle, as illustrated in Fig. \ref{fig:concept}.iv. This process can be repeated for multiple observed $\Delta f$ values to generate multiple candidate ellipses per sample which are all condensed into the same accumulation filter.

The next step is to identify the locations with the highest density of ellipse overlaps, and thus the locations most likely to signify an obstacle. These are identified by first setting the bottom 90\% of all nonzero bins to 0, as well as any bins that do not exceed a value of 1\% of the maximum value bin. All remaining bins are then checked to see if they are the maximum value bin within 50 bins (\qty{5}{m}), and are determined to be likely locations for obstacles if they meet these criteria.

The implementation is formally stated in Alg. \ref{alg:acoustic_mapping} and can also be seen in source code provided at \cite{github_anon}.

\begin{algorithm}[htbp]
\caption{Obstacle identification}
\label{alg:acoustic_mapping}
\SetAlgoLined
\DontPrintSemicolon

\SetKw{Ellipse}{Ellipse}

\KwIn{Discrete spectrogram $s(t)$, Robot poses $\mathbf{x}_S(t)$, $\mathbf{x}_L(t)$, Chirp parameters ($f_{\text{start}}$, $f_{\text{end}}$, $T$), Speed of sound $a$, Grid resolution $w$}
\KwOut{Set of obstacle highlight locations $H$}

\BlankLine
Initialize 2D accumulation grid $A(i, j) \leftarrow 0, \forall i, j$\;

\BlankLine
\For{each measurement step $t_k$}{
    Extract prominent frequency peaks $f_{\text{direct}}$ and $f_{\text{reflected},m}$ from spectrogram\;
    
    \For{each detected reflection peak $m$}{
        $\Delta f_m \leftarrow f_{\text{reflected},m} - f_{\text{direct}}$\;
        $\Delta d_m \leftarrow a \cdot \frac{\Delta f_m \cdot T}{f_{\text{start}} - f_{\text{end}}}$\;

        \BlankLine
        \For{each grid cell A(i,j)}{
            \If{\Ellipse{($\mathbf{x}_S(t)$, $\mathbf{x}_L(t), \Delta d_m$)} enters grid cell A(i,j)}{
                $A(i, j) \leftarrow A(i, j) + \text{intensity of } s(t) \text{ at } f_{\text{reflected},m}$
            }
        }
    }
}

\BlankLine
Apply thresholding: $[A(i,j) < P_{90}(A) \text{ or } A(i,j) < 0.01 \cdot \max(A)] \leftarrow 0$\;

Find all local maxima $H \in A(i,j)$ using a spatial radius window of $\lfloor 5 / w \rfloor$ bins\;

\BlankLine
\Return $H$\;
\end{algorithm}

\section{Simulation Results}

\begin{figure}
    \centering
    \includegraphics[width=\columnwidth]{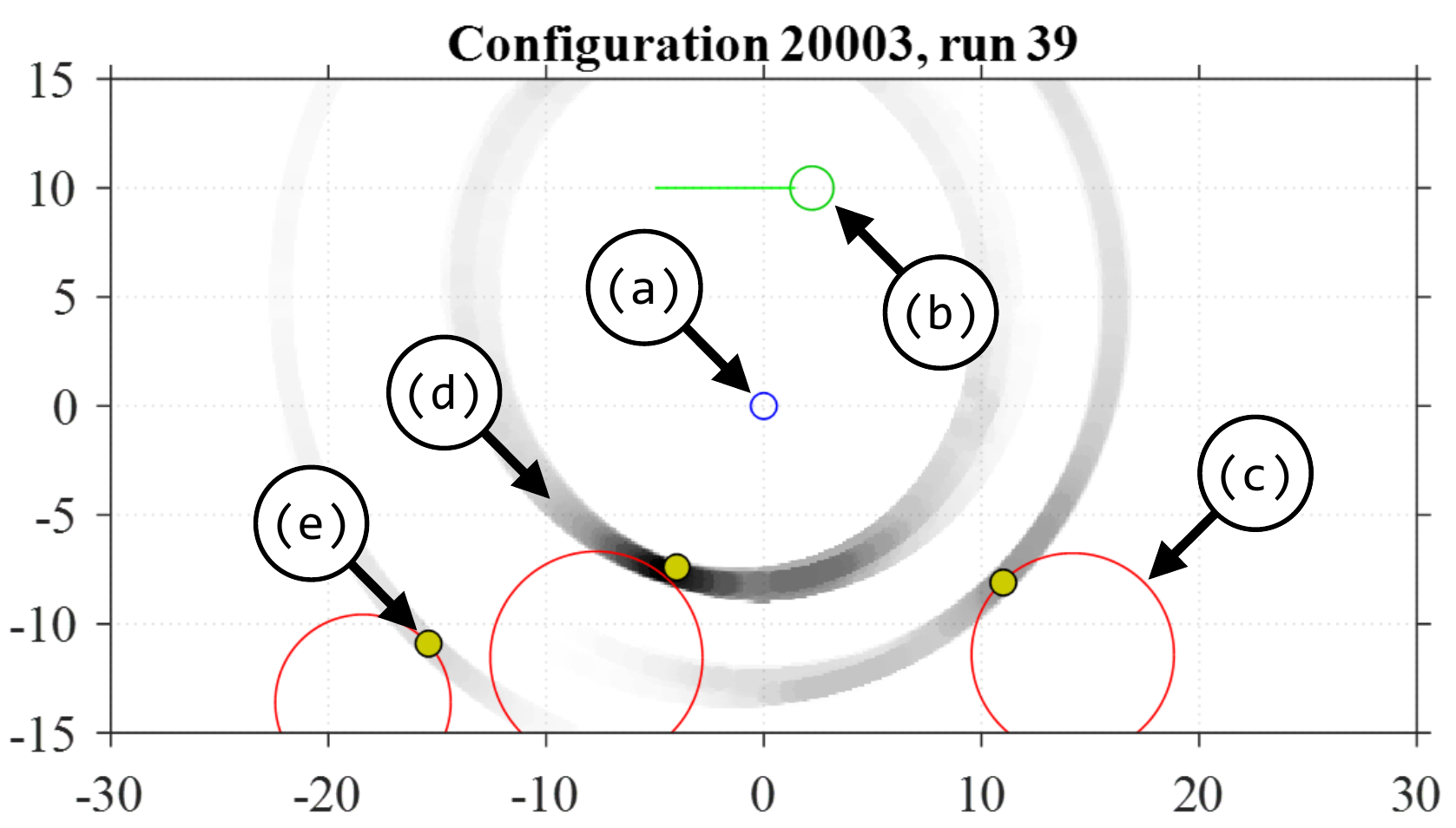}
    \caption{Simulation output showing the (a) source of the sound and the (b) listener; the green line represents the path traveled by the listener and the size of the listener is shown by the size of the green circle. (c) Three obstacles of varying sizes and positions are near the bottom of the simulated environment (red circles). (d) The accumulated ellipses are shown here, enlarged for clarity. (e) The darkest locations of the accumulations are identified as highlights (yellow circles).}
    \label{fig:sim_output}
\end{figure}

In this section we present the results of simulations and demonstrate the ability to use a \textbf{single chirp} to identify obstacles in the environment. Multiple chirps would enable refinements to obstacle detection so the results here represent a floor to the performance of our approach.  Fig. \ref{fig:sim_output} shows a post-processed simulation with highlights identified. In all tests, sound speed is set to \qty{343}{m/s}, and to match chirp duration and movement speed of a typical bird, chirp duration is set to \qty{0.5}{s}, and Listener speed is set to \qty{10}{m/s} (the approximate airspeed velocity of an unladen swallow). Obstacle diameters are randomized between \qtyrange{5}{10}{m}, and their positions are also randomized. Because the random seed is set at the beginning of every run, obstacle placements are consistent per-run across the configurations tested (i.e., run 1 has the same obstacles for configuration 1 as it does for run 1 of configuration 2). 

\subsection{Evaluation Metrics}

Our method functions by using overlapping ellipses to identify ``highlights,'' locations that are likely to be on or near the surface of an obstacle. Because of the many ways in which these ellipses can overlap and the possibility for false positives and false negatives, it is necessary to evaluate the accuracy of our method. We do this using two separate but related metrics:

{\texttt{score\_obstacle}} measures how well each obstacle is described by a highlight, and can be thought of as a measure of how well the method avoids false negatives. This measure is calculated by first assigning each highlight to the nearest obstacle, then identifying, per obstacle, the distance to the closest highlight assigned to it. The obstacle then receives a score of one divided by this distance, with a maximum score of one.

{\texttt{score\_highlight}} measures how well each highlight describes an obstacle, and can be thought of as a measure of how well the method avoids false positives. This score is calculated by determining the distance between each highlight and the closest obstacle to it. The highlight then receives a score of one divided by this distance, with a maximum score of one.

\subsection{Simulation Results for a Single Listener}

\begin{table}[]
\centering
\vspace{1.5mm}
\caption{Mean Results for a single listener}
\label{tab:single_results}
\resizebox{\columnwidth}{!}{%
\includegraphics[]{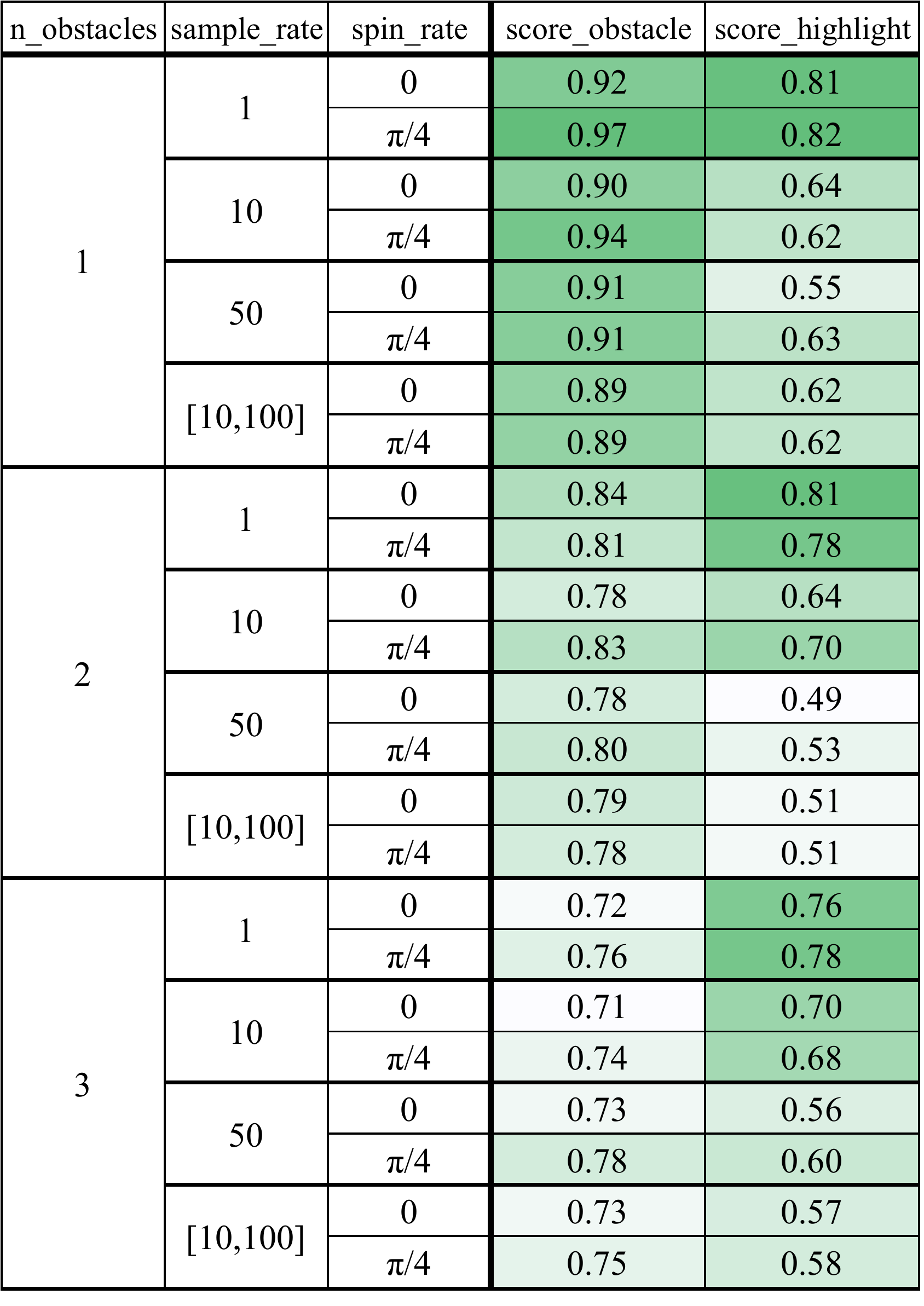}
}
\end{table}

To validate our approach, we sought to quantify the effect of various parameters on the ability to accurately identify obstacles. Specifically, we varied the following:

\texttt{n\_obstacles}, the number of obstacles present in the environment. In all cases, these obstacles are placed on the opposite side of the Source as the Listener. We consider the cases of environments with 1, 2, and 3 obstacles. No attempt was made to prevent one obstacle obscuring another beyond preventing overlaps, which can sometimes prevent an obstacle from being detected. Varying this enables us to understand how well the approach will work in more complex environments.

\texttt{spin\_rate}, the rotational speed of the Listener that causes the Listener to move in an arc rather than in a straight line. Here, we consider two cases: one in which the {spin\_rate} is 0 (the Listener travels in a straight line), and one in which the {spin\_rate} is ${\pi/4}$ rad/s, arcing away from the Listener. Varying this enables us to understand whether the path that the Listener takes affects the ability to identify obstacles.

\texttt{sample\_rate}, the number of milliseconds between checking the heard frequencies and calculating the ellipses; every simulation tick represents \qty{1}{ms}. We consider three constant periods of 1, 10, and \qty{50}{ms}. Additionally, we also consider a non-constant case where each new measurement and recalculation occurs \qtyrange{10}{100}{ms} (uniformly distributed random number) after the previous measurement. This represents a case where a robot needs to dynamically adjust where it focuses its computational resources. Varying this enables us to understand how important it is to have frequent measurements, as well as to demonstrate how the system can function when the robot cannot constantly monitor its sensors, a key benefit relative to the more traditional ``click'' approach.

We considered all combinations of the listed configuration parameters, a total of 24 different configurations. We ran each configuration 100 times and calculated the mean \texttt{score\_obstacle} and \texttt{score\_highlight} for each configuration. The results of our investigation are summarized in Table \ref{tab:single_results}. 

Several general trends can be readily observed in this table. The first is that as the number of obstacles increases, the quality of the mapping decreases. Because the \texttt{score\_highlight} is far less affected, this is likely due to an increasing number of obstacles making it more likely that one obstacle obscures another. The second, is that the method is robust against different sampling rates, including when using random and inconsistent sampling. Finally, an arcing motion by the Listener appears to slightly improve the ability to identify obstacles. The exact mechanism for this improvement is unclear, but is likely related to the motion resulting in a more varied set of candidate ellipses.

\subsection{Simulation Results for Two Listeners}

\begin{figure}
    \centering
    \vspace{1.5mm}
    \includegraphics[trim=2mm 3mm 3mm 2mm, clip,width=0.8\linewidth]{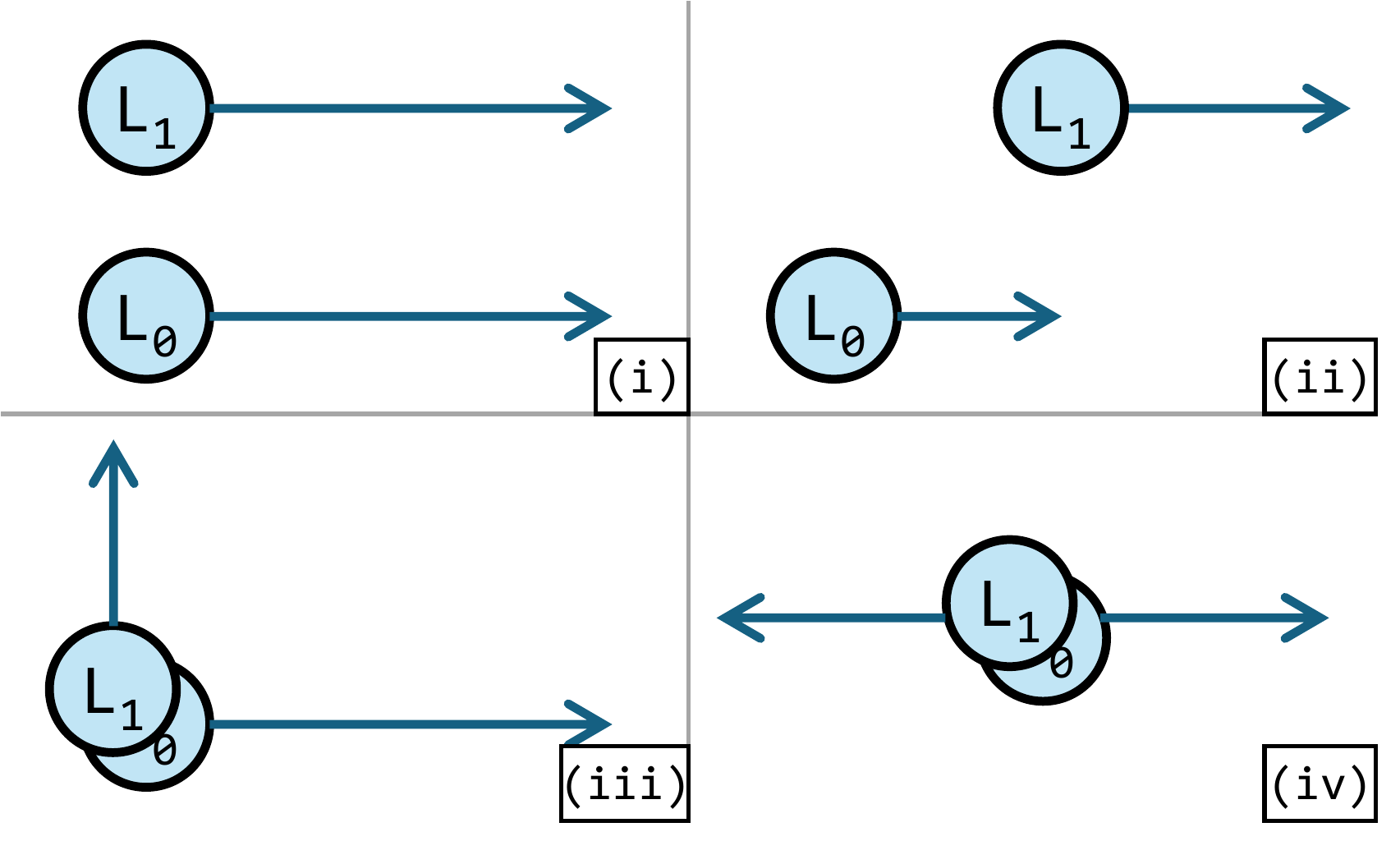}
    \caption{Two-listener simulations consider four distinct movement patterns: (i) Same direction and x start position, offset by 5 units on the y axis; (ii) Same direction but with L1's x start position set to L0's x end position, offset by 5 units on the y axis, mirrored x starting location; (iii) same starting location but perpendicular movement; (iv) same centered starting location, opposite movement.}
    \label{fig:movement_patterns}
\end{figure}

\begin{table}[]
\centering
\vspace{1.5mm}
\caption{Mean Results for two listeners.}
\label{tab:multiple_results}
\resizebox{\columnwidth}{!}{%
\includegraphics[]{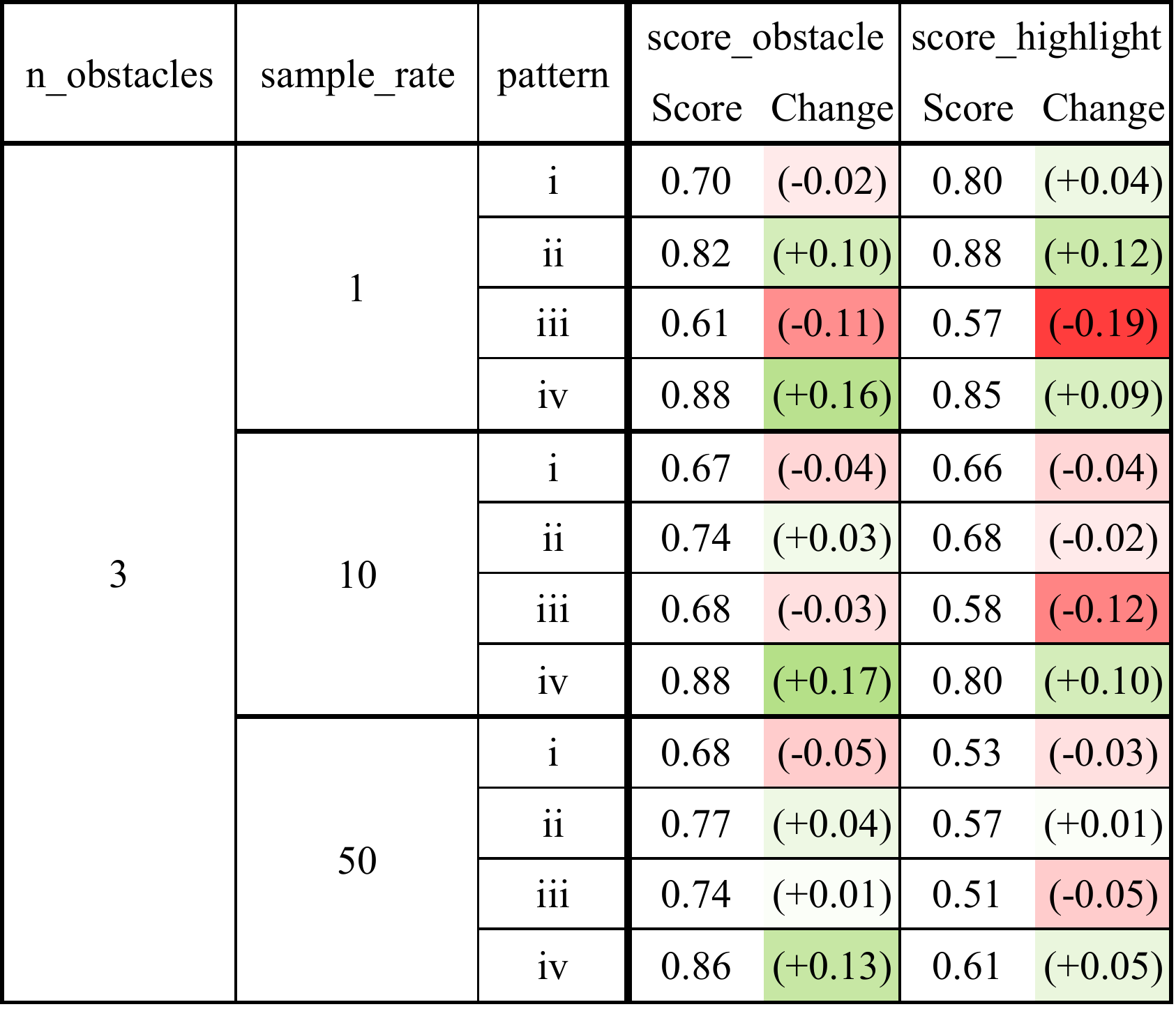}
}
\end{table}

In this subsection we consider a case where two listeners are moving and are cooperating on the same shared map, specifically we consider the four movement patterns shown in Fig. \ref{fig:movement_patterns} to investigate how they impact the \texttt{score\_obstacle} and \texttt{score\_highlight} metrics. To narrow the scope of this investigation, for the results presented here, we only consider the case of $\texttt{n\_obstacles} = 3$, $\texttt{spin\_rate} = 0$, and $\texttt{sample\_rate} = 50$ since this represented a configuration that proved to be challenging for a two-listener system. We ran each configuration 60 times.

As summarized in Table \ref{tab:multiple_results}, what we found was that different movement patterns resulted in different impacts to the ability for the system to accurately identify obstacles. Generally, movement pattern (i) had a minor impact on obstacle identification, with a negative impact for low sample rates; movement pattern (ii) seemed to generally benefit the system for higher sample rates with minimal impact to slower sample rates; (iii) appears to be harmful to the ability for the system to function; whereas (iv) yielded very positive benefits. In other words, the system appeared to benefit from having the greater coverage on the x axis (ii, iv), and adding a listener with a very similar movement pattern (i) did not add useful information. The effects of movement pattern (iii) are more difficult to understand and explain, but appear to be the result of the perpendicular motion causing the ellipses to accumulate in such a way that the local maxima would no longer always lie on the obstacles. In preliminary tests we found that movement along the y axis did not result in as good of results as movement along the x axis, and so it may also simply be the result of movement pattern (iii) adding lower-quality data to a system that already has fairly good quality data with the single listener moving along the x axis.

\section{Experimental Evaluation}

\subsection{Experimental Process}

\begin{figure}
    \centering
    \vspace{1mm}
    \includegraphics[trim=0mm 0mm 0mm 50mm, clip, width=1.0\linewidth]{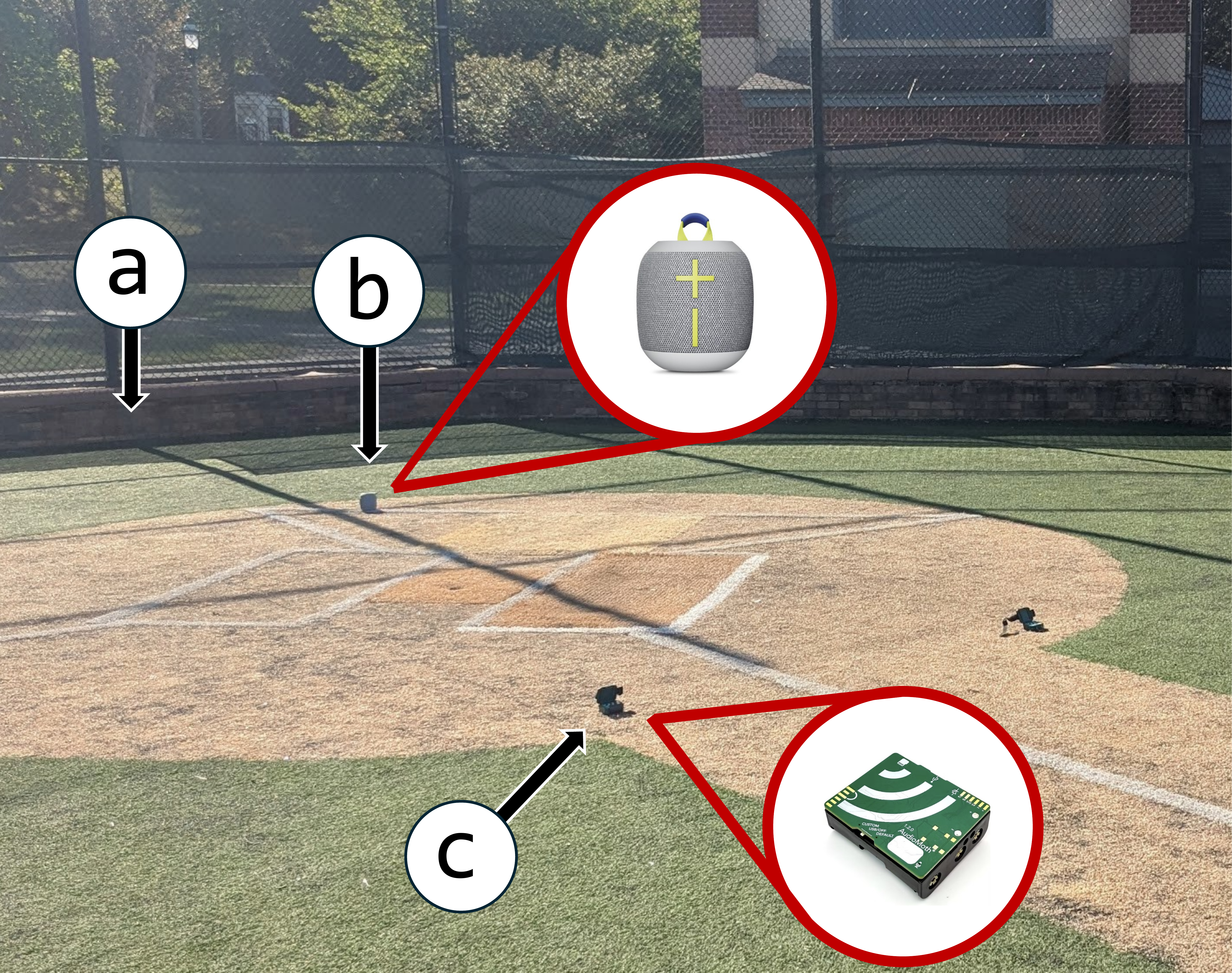}
    \caption{Photograph of the experimental setup for Environment A, in a {baseball} field of {Branch Brook Park} in {Newark, NJ}. The following elements were considered to mirror the simulation: (a) Wall to Detect, (b) WONDERBOOM 4 Portable Bluetooth Speaker, (c) AudioMoth Acoustic Logger.}
    \label{fig:diamond}
\end{figure}

\begin{figure}
    \centering
    \includegraphics[trim=1mm 1mm 1mm 1mm, clip, width=1.0\columnwidth]{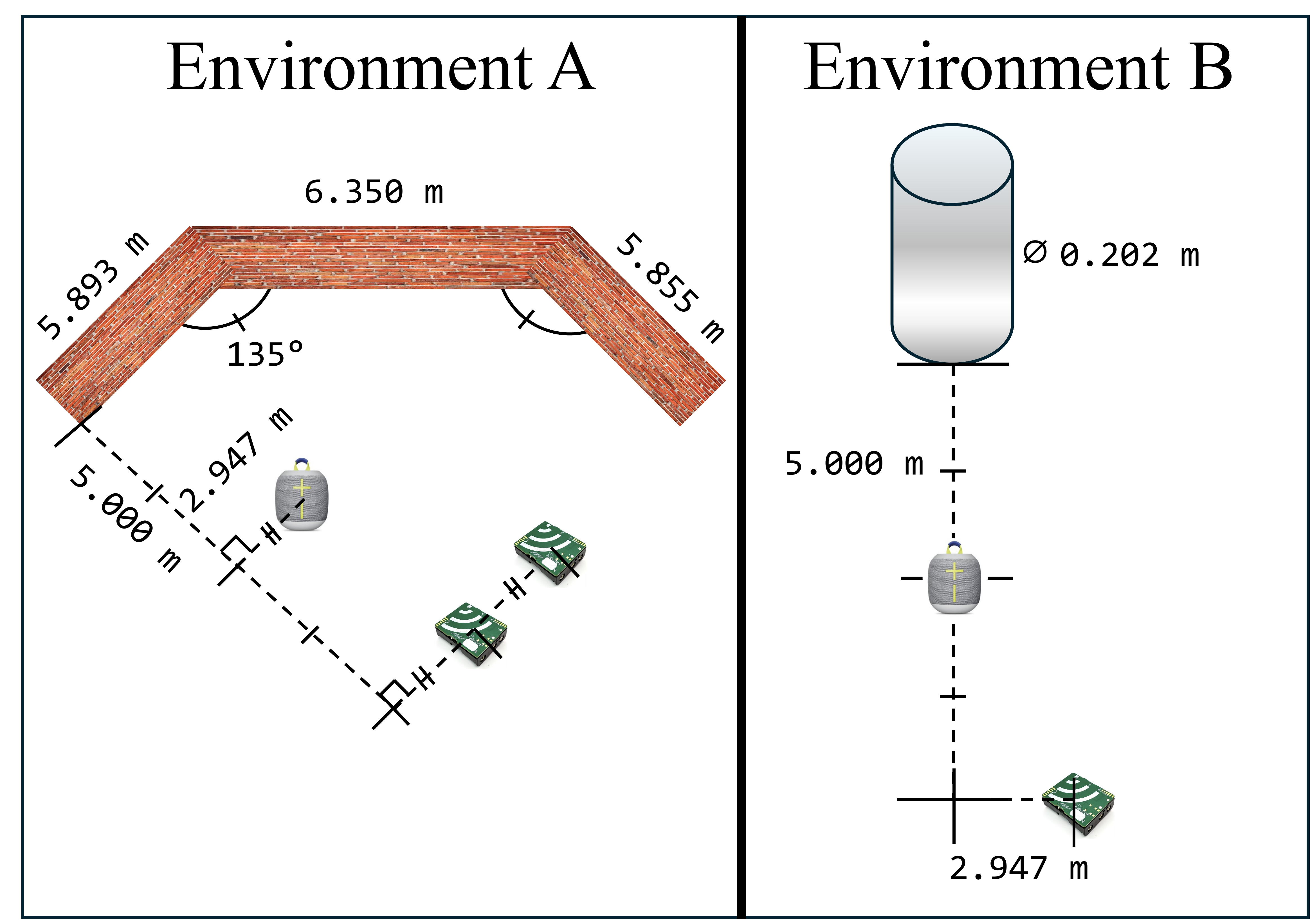}
    \caption{Diagrams of the environment configurations used for the field experiment. (a) Environment A is located in a baseball field with a brick wall, one WONDERBOOM 4 Bluetooth speaker, and two AudioMoth microphones. (b) Environment B is located in an open grass field with a steel pole, one speaker, and one microphone.}
    \label{fig:ex_setup}
\end{figure}

To validate the real-world viability of our distance detection algorithm, we conducted field experiments in {Branch Brook} Park in {Newark, NJ}. AudioMoth Acoustic Loggers and a WONDERBOOM 4 Portable Bluetooth speaker were used to model the simulation setup ({Fig. \ref{fig:diamond}), with distributed listeners in fixed positions representing a linear flight path for simplification. The internal real-time clocks of the AudioMoth loggers were synchronized to UTC using the AudioMoth app. Two different environment configurations were considered with regard to location and listener positioning, as seen in {Fig. \ref{fig:ex_setup}}. The speaker emits a linear frequency sweep with a 48 kHz sample rate from 10 kHz to 0.2 kHz for which the agents in the system know both the magnitude and the timing. Three different chirp lengths were used: 1 s, 0.5 s, and 0.25 s. Three trials were conducted for each environment configuration, with 50 chirp repetitions and a one second gap between chirps.

A high-pass filter of 100 Hz was applied to the collected data to reduce noise. Additionally, a Savitzky-Golay filter was applied to smooth the data while preserving the frequency peaks. The processing mirrors that of the simulation, using FFT to translate the signals into their constituent frequencies. Chirps are processed individually and the median value is considered for each environment configuration.

\begin{figure}
    \centering
    \includegraphics[width=1\linewidth]{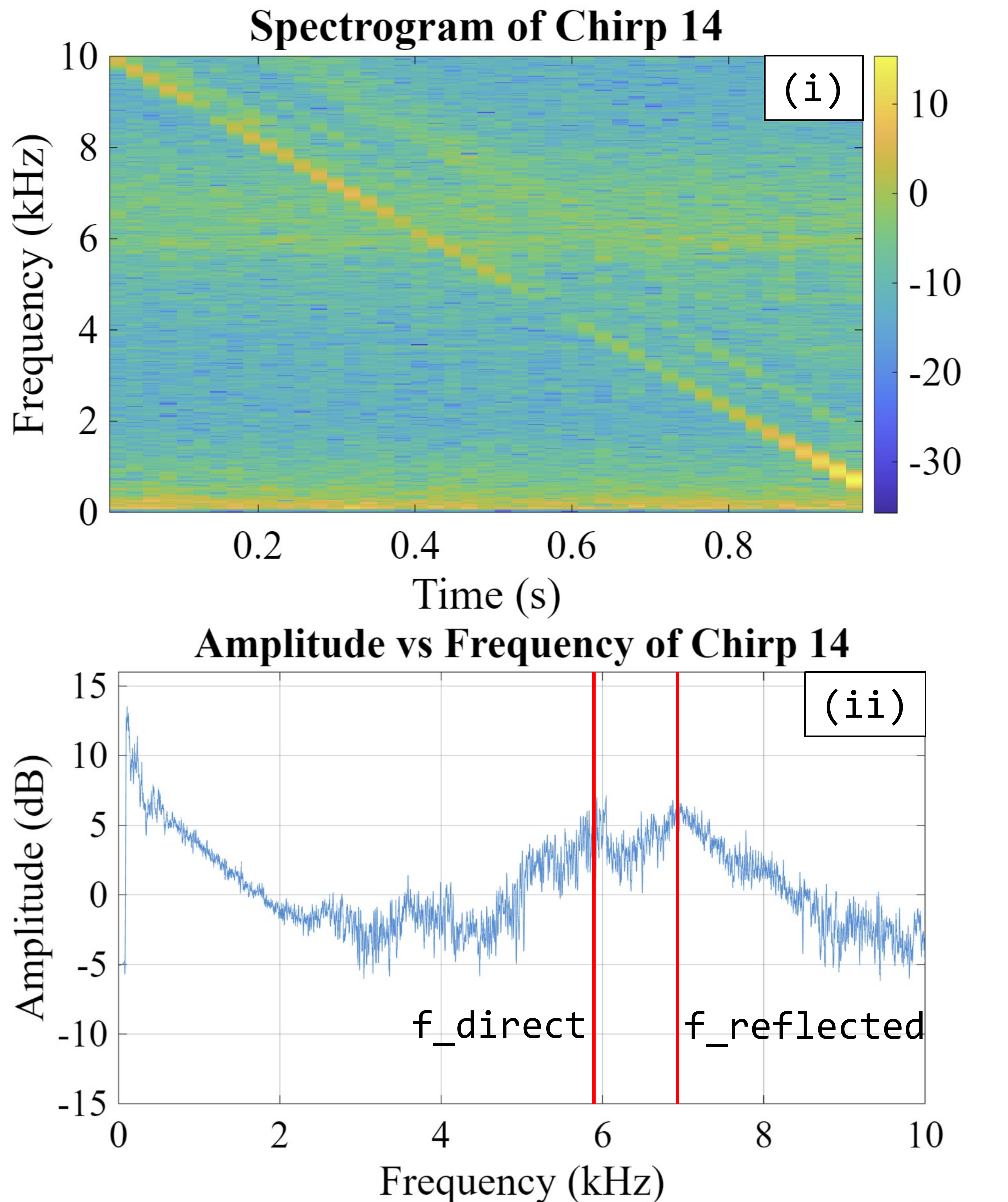}
    \caption{Spectrogram (i) and Amplitude vs. Frequency plot (ii) for Listener 2 in Environment B with a one second chirp duration, Run 2 of 3, Chirp 14 of 50.}
    \label{fig:freqplot}
\end{figure}

\subsection{Experimental Results}

\begin{figure}
    \centering
    \includegraphics[width=1\columnwidth]{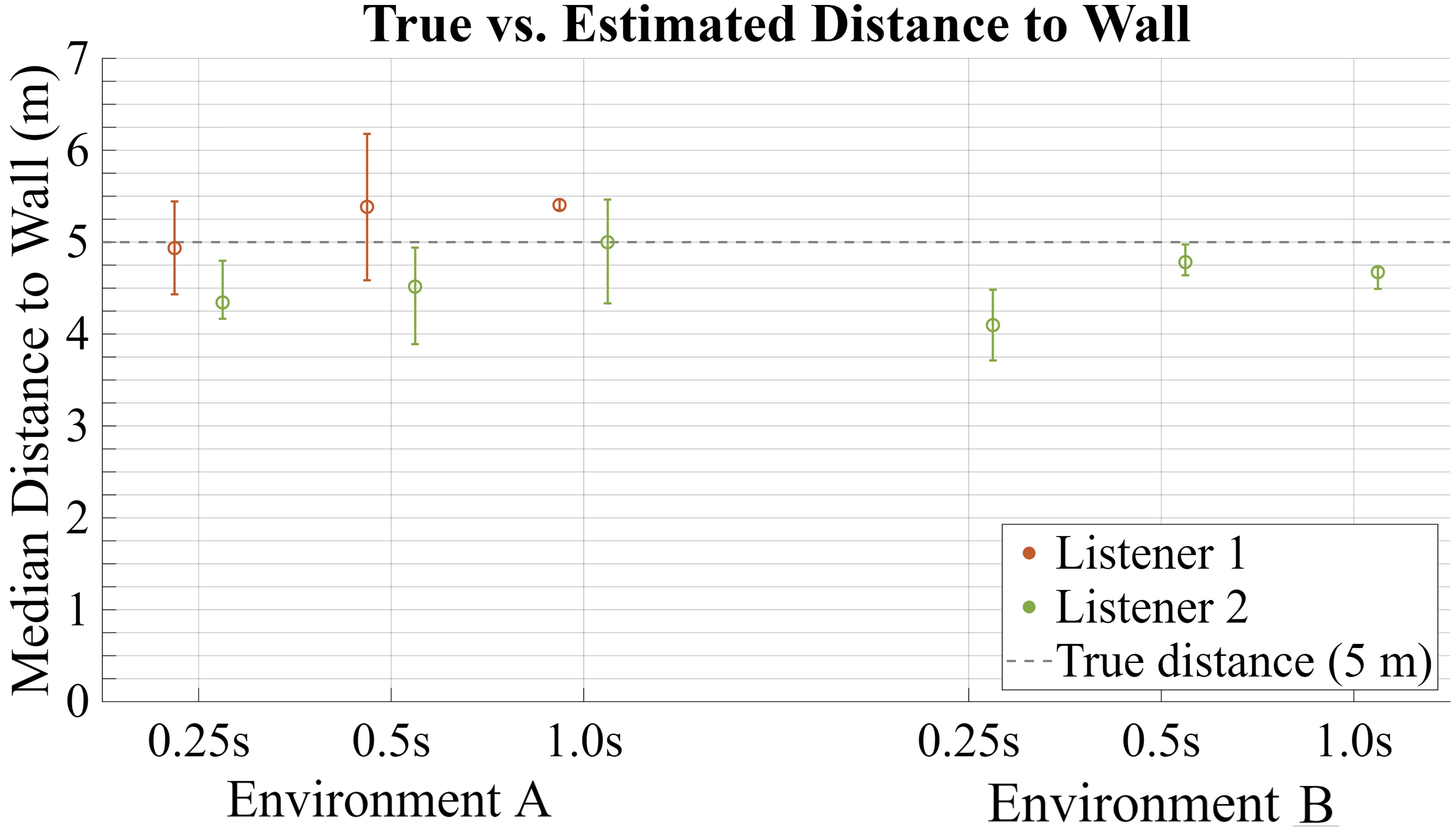}
    \caption{Median wall distance estimates across 150 chirps per environment configuration. Error bars represent the 95\% confidence interval.}
    \label{fig:prelim_res}
\end{figure}

{Fig. \ref{fig:freqplot}} demonstrates our signal processing techniques used on a real chirp in the field experiments, which clearly identifies the frequency peak separation between the sound traveling directly from the source to the listener, and the sound traveling from the source to the obstacle to the listener.

{Fig. \ref{fig:prelim_res}} shows that the preliminary results of our approach are quite promising. The percent error for each listener in all the environment configurations ranged from -0.01\% (Listener 2, Environment A, \qty{1.0}{s} chirp) to -18\% (Listener 2, Environment B, \qty{0.25}{s} chirp). As expected, longer chirp durations generally improved the accuracy of the distance estimation due to reduced noise and sharper frequency peaks. Interestingly,  the difference in the wall distance estimates with different chirp lengths is minimal, suggesting that a variety of frequency sweeps of known timing and magnitude may be viable using our method of range finding. Our preliminary results successfully demonstrate the viability of our range finding approach, which sets the foundation for future work to implement in a real-time system.

{\section{Discussion and Limitations}}

The simulation and experimental results establish the promise of the presented approach to multi-agent acoustic mapping. As summarized in Table \ref{tab:sensor_comparison}, the approach has several fundamental advantages compared to more traditional ranging approaches. However, the exploratory nature of this paper means that there is still a need to evaluate the method at higher fidelity. 

The 2D particle-based simulation is sufficient for demonstrating the core logic of the frequency-gap-to-spatial-map pipeline, it explicitly ignores complex real-world acoustic phenomena. One such phenomenon is Doppler shift, which is supported by the simulation but which was purposefully omitted to enable a more ``pure'' exploration of the basic concept. In preliminary investigations, the Doppler shift had the effect of simply shifting the estimated locations in an apparently consistent and predictable manner, suggesting that the effects of the Doppler shift could be reversed with some manner of realignment skew. Further, including Doppler information could unlock sensing improvements by adopting Doppler techniques used by  many radar and sonar systems to refine mapping capabilities. 

A notable feature of the system is that it is inherently robust to many unmodeled intricacies due to its continuous nature. For example, the fact that the listener is constantly moving makes it unlikely that multipath reflections will be present across the entire chirp; as demonstrated in Table \ref{tab:single_results}, the system is robust to sporadic sensor readings. The continuous nature of the approach also means that exact synchronization between the speaker and listener is not necessary since the exact location of the listener at each time is not particularly important and neither is the timing of the measurement. This decoupling of measurement timing provides a significant architectural advantage for large and low-cost systems. Further, although the presented method assumes shared knowledge of chirp duration and frequency range, the listener inherently receives all information needed determine these parameters via the chirp signal itself; any such corrective step would also serve to self-calibrate to different clock speeds, speaker characteristics, and microphone response.

The most limiting aspect of the presented approach is likely the accumulation filter approach to translating reflection ellipses into likely obstacle locations. It is unclear how well this approach will scale to more complex real-world environments or how to adapt the system towards synthesizing multiple chirp events. A likely approach would be to introduce a formalized probabilistic approach that considers conditional probabilities. This would also be useful towards accounting for and mitigating the effects of localization error.

\begin{table}[t]
\caption{Qualitative comparison to established ranging methods}
\label{tab:sensor_comparison}
\centering
\begin{tabular}{p{0.4in} p{0.2in} p{1.1in} p{1in}}
\hline
\textbf{Method} & \textbf{Cost} & \textbf{Update Rate } & \textbf{Multi-Agent Interference} \\ \hline
\textbf{Ultrasonic ToF} & Low & \textbf{Low}: Limited to round-trip echo time. & \textbf{High}: Overlapping unmodulated pulses cause crosstalk. \\ \hline
\textbf{Standard Sonar} & Med. & \textbf{High}: Continuous beat frequencies. & \textbf{Medium}: Requires distinct signal bands. \\ \hline
\textbf{2D \newline LiDAR} & High & \textbf{High}: Depends on mechanical rotation. & \textbf{Low}: Typically not an issue.\\ \hline
\textbf{Acoustic Ellipse (Ours)} & \textbf{Low} & \textbf{High}: Measurement timing decoupled from signal emission. & \textbf{Low}: Most agents will be passive listeners.\\ \hline
\end{tabular}
\end{table}

\section{Conclusion}
Our work establishes the efficacy of cooperative acoustic mapping using frequency-modulated signals. Simulation results showed that the presented method is robust to low and inconsistent sampling rates, a key advantage for compute-constrained robots. We further demonstrated that a cooperative approach can dramatically enhance mapping accuracy, and showed that core aspects of the system can be easily implemented in real-world environments.

There are several avenues of research we plan to pursue in future work, foremost of which is the unifying of our experimental approach with the algorithm in a real-time, closed-loop autonomous system. We also plan to investigate machine-learning approaches to signal processing to identify potential improvements to the ellipse-accumulation approach we presented, including an extension to 3D. Finally, the presented system lacks consideration for measurement uncertainties and synthesizing results from multiple chirp events, which is likely to aid in the observed degradation in more complex environments.

\section*{ACKNOWLEDGMENT}

\textit{AI Disclosure}: Gemini 2.0 Fast was used to create several of the helper functions used for the simulator. All such functions make specific note of AI use in the comments of the source code. These functions were generated by defining specific input parameters and specific outputs; all such functions were validated for correctness. 

\balance


\end{document}